%% file: main.tex
\documentclass[runningheads]{llncs}

\usepackage{eccv}

\usepackage{xcolor}  
\usepackage{booktabs,tabularx,caption}
\usepackage{multirow}
\usepackage{soul}
\usepackage{comment}
\usepackage{mathrsfs}
\usepackage{textcomp}
\usepackage{bbm}
\usepackage{verbatim}
\usepackage[accsupp]{axessibility}
\usepackage{wrapfig}
\usepackage{times}
\usepackage{epsfig}
\usepackage{graphicx}
\usepackage{eccvabbrv}
\usepackage[export]{adjustbox}
\usepackage{graphicx}
\usepackage{booktabs}
\usepackage{float}

\usepackage[accsupp]{axessibility}  % Improves PDF readability for those with disabilities.
\input{ourcommands}

\usepackage{hyperref}

\usepackage{orcidlink}

\begin{document}

\title{Active Coarse-to-Fine Segmentation of Moveable Parts from Real Images} 


\author{Ruiqi Wang\inst{1}\orcidlink{0009-0000-3379-6103} \and
Akshay Gadi Patil\inst{1}\orcidlink{0000-0003-1429-3804} \and
Fenggen Yu\inst{1}\orcidlink{0000-0003-1591-4668}\and
Hao Zhang\inst{1,2}\orcidlink{0000-0003-1991-119X}}

\authorrunning{Ruiqi Wang et al.}

\institute{Simon Fraser University, Burnaby, Canada\\
\email{\{ruiqi\_w,agadipat,fenggen\_yu,haoz\}@sfu.ca} 
\and 
Amazon
\vspace{-10pt}}

\maketitle

\begin{center}
     \centering
     % \captionsetup{type=figure}
     \includegraphics[width=\textwidth]{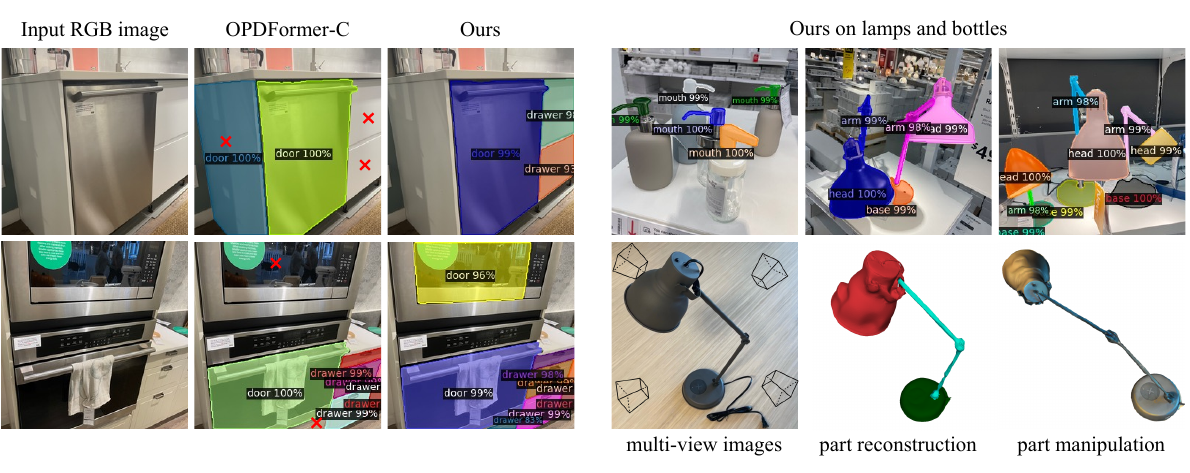}
     \vspace{-10pt}
     \captionof{figure}{\rzz{Our} instance segmentation of \xeable parts, with semantic labels, on real-world photos. Comparison is made with OPDFormer-C \rzz{(OPD = {\em openable\/} part detection)}, the current state of the art, where small red \textcolor{red}{\textbf{$\times$}}s indicate \rz{erroneous or missed labels.} \chk{Our method generalizes to {\em non-openable\/} parts, e.g., on lamps and bottles (top right).}
     As an application of accurate \xeable part segmentation, we can manipulate 3D reconstructions of articulated objects (bottom right). } % and more details in Section~\ref{sec:app}.}
     \label{fig:teaser}
\end{center}%
 \input{sec/0_abstract}

 \input{sec/1_intro}
 \input{sec/2_related_works}
 \input{sec/3_problem_statement}

 \input{sec/4_method}

 \input{sec/5_datasets}

 \input{sec/6_results}

 \input{sec/7_application}
 \input{sec/8_conclusion}
\clearpage
\section*{Acknowledgements}
This work was supported in part by NSERC. We thank all anonymous reviewers and area chairs for their valuable comments, Hanxiao Jiang, Hang Zhou for insightful discussion, and Mingrui Zhao for help with data annotation.
\bibliographystyle{splncs04}
\bibliography{main}
\end{document}

%% file: ourcommands.tex
\definecolor{green}{rgb}{0, 0.5, 0}
\definecolor{orange}{rgb}{0.8, 0.6, 0.2}
\definecolor{red}{rgb}{1.0, 0.0, 0.0}
\definecolor{teal}{rgb}{0.0, 0.4, 0.4}
\definecolor{purple}{rgb}{0.65,0,0.65}
\definecolor{saffron}{rgb}{0.95,0.75,0.2}
\definecolor{turquoise}{rgb}{0.0,0.5,0.5}
\definecolor{black}{rgb}{0.0, 0.0, 0.0}
\definecolor{gray}{rgb}{0.5, 0.5, 0.5}
\definecolor{blue}{rgb}{0.0, 0.0, 1.0}

\newcommand{\rqw}[1]{{\color{black}#1}}

\newcommand{\agp}[1]{{\color{black}#1}}
\newcommand{\rz}[1]{{\color{black}#1}}
\newcommand{\rzz}[1]{{\color{black}#1}}
\newcommand{\chk}[1]{{\color{black}#1}}

\newcommand{\xeable}{{\color{black}moveable~}}

%% file: sec/0_abstract.tex
%%%%%%%%% ABSTRACT
\begin{abstract}
We introduce the first {\em active learning\/} (AL) model for high-accuracy instance segmentation of \xeable parts from RGB images of {\em real indoor scenes}.
\rzz{Specifically, our goal is to obtain {\em fully validated\/} segmentation results by humans while {\em minimizing manual effort\/}.} To this end, we employ a transformer that utilizes a masked-attention mechanism to supervise the active segmentation. 
To enhance the network tailored to \xeable parts, we introduce a {\em coarse-to-fine\/} AL approach which first uses an {\em object-aware\/} masked attention and then a {\em pose-aware\/} one, \rz{leveraging the hierarchical nature of the problem and a correlation between \xeable parts and object poses and interaction directions.}
\rzz{When applying our AL model to 2,000 real images, we obtain fully validated \xeable part segmentations with semantic labels,}
\rzz{by only needing to manually annotate \rqw{11.45\%} of the images. This translates to significant (60\%) time saving over manual effort required by the best non-AL model to attain the same segmentation accuracy.}
At last, we contribute a dataset of \rqw{2,550} real images with annotated \xeable parts, demonstrating its superior quality and diversity over \rzz{the best alternatives.}
\end{abstract}

%% file: sec/1_intro.tex
%%%%%%%%% BODY TEXT
\section{Introduction}
\label{sec:intrp}

Most objects we interact with in our daily lives have dynamic movable parts, where the part movements reflect how the objects function.
Perceptually, acquiring a visual and actionable understanding of object functionality is a fundamental task.
In recent years, motion perception and functional understanding of articulated objects have received increasing attention in vision, robotics, and VR/AR. %As a very common component of the indoor scene in both synthetic and natural environments, articulated objects involve different human interactions and manipulations. 
Aside from per-pixel or per-point motion prediction, %\chk{2D} 
\rzz{the detection and segmentation of \xeable parts plays a vital role in embodied AI applications involving robot manipulation and action planning.}

% serves as the basis for many downstream tasks, including robot manipulation, action planning, and part-level 3D reconstruction. 
\begin{figure}[tb]
\centering
  \includegraphics[width=0.99\textwidth]{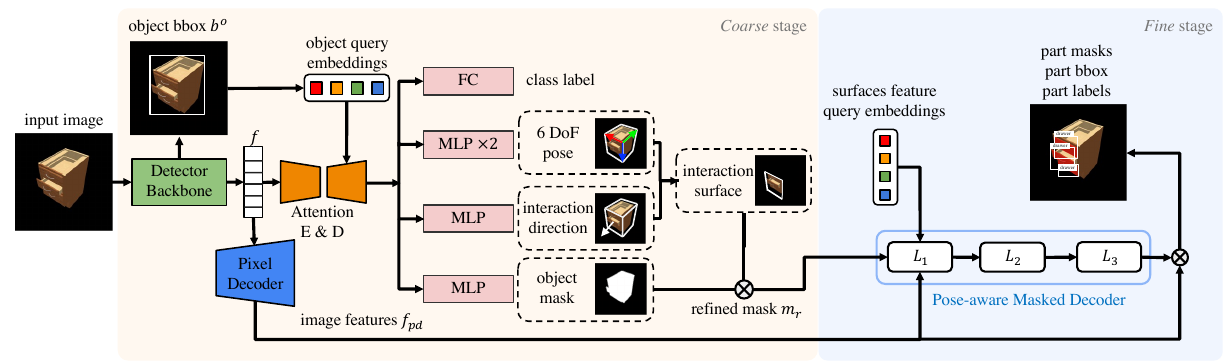}
  \caption{\rqw{Overview of our pose-aware masked attention network for \xeable part segmentation of articulated objects in real scene images. Utilizing a two-stage framework, we first derive a \emph{coarse} segmentation by predicting the object mask, its 6 DoF pose, and the interaction direction, subsequently isolating the interaction surface of the objects. In the \emph{fine} segmentation stage, we combine the object mask and interaction surface to form a refined mask, enabling the extraction of fine-grained instance segmentation of \xeable parts.} }
  \label{fig:method}
  \vspace{-10 pt}
\end{figure}

In this paper, we tackle the problem of {\em instance\/} segmentation of \xeable parts \rz{in one or more articulated objects} from RGB {\em images of real indoor scenes\/}, \rz{as shown in Figure~\ref{fig:teaser}.}
\rz{Note that we use the term articulated objects in a somewhat loose sense to refer to all objects whose parts can undergo motions; such motions can include
opening a cabinet door, pulling a drawer\footnote{Strictly speaking, articulations are realized by ``two or more sections connected by a {\em flexible\/} joint,'' which would not
include drawer sliding. However, as has been done in other works in vision and robotics, we use the term loosely to encompass more general part motions.}, and the movements of a
lamp arm.}
Most prior works on \rz{motion-related} segmentations~\cite{yan2020rpm,li2020category,huang2021multibodysync} operate on point clouds, which are more
expensive to capture than images while having lower resolution, noise, and outliers.
\rz{Latest advances on large language models (LLMs) \rzz{and vision-language models (VLMs)} have led to the development of powerful generic models such as SAM~\cite{kirillov2023segany}, which can
excel at generating quality object masks and exhibit robust zero-shot performance across diverse tasks owing to their extensive training data. However, these methods 
remain limited in comprehensive understanding of \xeable object parts.} % to achieve high part segmentation accuracy.}
% Also, point samples acquired in real-world setting are often tempered with noise, missing data, and outliers. 

To our knowledge, OPD~\cite{jiang2022opd}, for ``openable part detection", \rz{and its follow-up, OPDMulti~\cite{sun2024opdm}, for ``openable part detection for multiple objects'', represent} 
the state of the art in \xeable part segmentation from images. \rz{However, despite the fact that both methods were trained on real object/scene images, there still remains a 
large gap between synthetic and real test performances: roughly 75\% vs.~30\% in segmentation accuracy~\cite{sun2024opdm}. The main reason is that manual instance segmentation on real images to form ground-truth training data is too costly. As a remedy, OPD and OPDMulti both opted to manually annotate 3D mesh or RGB-D {\em reconstructions\/} from real-world articulated object scans and project the obtained segmentation
masks \rzz{to 2D.} % the real images. 
\rzz{Thus, for each reconstructed 3D scene, there is only a one-time annotation, in 3D, required, after which thousands of annotated images can be rendered.} Clearly, such an indirect annotation still leaves a gap between rendered images of digitally reconstructed 3D models and real photographs, with both reconstruction errors and 
re-projection errors due to view discrepancies hindering the annotation quality on images. }

\rz{To close the aforementioned gap by addressing the annotation challenge}, we present an {\em active learning\/} (AL)~\cite{AL_survey2014,AL_survey2020,AL_comp_survey2022} 
approach to obtain high-accuracy instance segmentation of \xeable parts, \rz{with semantic labels, {\em directly on\/}} real scene images \rz{containing one or more articulated objects.}
AL is a semi-supervised learning paradigm, relying on human feedback to continually improve the performance of a \rzz{learned} segmentation model. 
%As with most human-in-the-loop approaches, the key criterion for success in AL is to minimize human effort. 
\rzz{Specifically, our goal in this work is to obtain {\em fully validated\/} segmentation results by humans while {\em minimizing manual segmentation efforts.} In other words, we would like the human to manually segment as few images as possible while ensuring that {\em all\/} the images in our dataset have been segmented accurately, either by a neural segmentation network that is trained by available ground-truth data or by human.}
To this end, we employ a transformer-based~\cite{dosovitskiy2020image} segmentation network that utilizes a masked-attention
mechanism~\cite{cheng2022masked}. To enhance the network for \xeable part segmentation, we introduce a {\em coarse-to-fine\/} AL model which first uses an
{\em object-aware\/} masked attention and then a {\em pose-aware\/} one, leveraging \rz{the hierarchical nature of the problem} and a correlation between \xeable parts and \rz{object poses
and interaction directions.} % and leading to improved handling of multiple articulated  objects in an image.

\rz{As shown in Figure~\ref{fig:method}, in the \emph{coarse} annotation stage, our AL model with object-aware attention predicts object masks, poses, and interaction directions, so as to help isolate 
interaction surfaces on the articulated objects. In the \emph{fine} annotation stage, we combine the object masks and interaction surfaces to predict refined segmentation masks for
\xeable object parts, also with human-in-the-loop.} Unlike prior works on active segmentation~\cite{xie2022towards,tang2022active} which mainly focused on the efficiency of human annotations \rzz{using point- or region-based supervision for fast labeling, we optimize the human-in-the-loop pipeline to reduce AL iterations and samples required for manual annotation.} Our network learns the regions-of-interests (ROIs) from the pose-aware masked-attention decoder for better segmentation sampling in AL iterations \rz{in the second stage, \rzz{where we categorize samples in different branches for further training, testing, and annotation.}}

In summary, our main contributions include:

\begin{itemize}
\vspace{-3pt}
\item We introduce the first AL framework for instance segmentation of \xeable parts from RGB images of real indoor scenes. 
\rzz{When applying our AL model to 2,000 real images, we obtain fully validated \xeable part 
segmentations with semantic labels, by only needing to manually annotate 11.45\% of
the images. This translates to significant (60\%) time saving over manual effort
required by the best non-AL model, i.e., OPDFormer-C~\cite{sun2024opdm}, to attain the same segmentation accuracy.}

\item 
Our coarse-to-fine AL model, with both object- and pose-aware masked-attention mechanisms, lead to reduced human effort and improved accuracy in \xeable part segmentation over state-of-the-art (SOTA) methods: OPD~\cite{jiang2022opd} \rz{and OPDMulti~\cite{sun2024opdm}}. 

\item Our \rz{scalable} AL model allows us to accurately annotate a dataset of \rqw{2,550} real photos of articulated objects in indoor scenes. We show the superior quality and diversity of our new dataset over \rz{current alternatives~\cite{jiang2022opd,sun2024opdm}}, and the resulting improvements in segmentation accuracy. 
\end{itemize}

%% file: sec/2_related_works.tex
\section{Related Works}
\label{sec:rw}
% \rqw{
% \paragraph{Parts segmentation of articulated objects.}

% \paragraph{Active learning for image segmentation.}

% \paragraph{}
% }

\agp{
\paragraph{\textbf{Articulated objects dataset.}}
The last few years have seen the development of articulation datasets on 3D shapes. Of the many, ICON \cite{hu2017learning} build a dataset of 368 moving joints corresponding to various parts of 3D shapes from the ShapeNet dataset~\cite{chang2015shapenet}. The Shape2Motion dataset \cite{wang2019shape2motion} provides kinematic motions for 2,240 3D objects across 45 categories sourced from ShapeNet and 3D Warehouse~\cite{trimble_warehouse}. The PartNet-Mobility dataset~\cite{Xiang_2020_SAPIEN} consists of 2,374 3D objects across 47 categories from the PartNet dataset~\cite{Mo_2019_CVPR}, providing motion annotations and part segmentation in 3D. 

All these datasets are obtained via manual annotations and are \emph{synthetic} in nature. Since sufficient training data is made available by these synthetic datasets, models trained on them can be used for fine-tuning on \emph{real-world} 3D articulated object datasets with limited annotations. \chk{However, models trained exclusively on \emph{synthetic} data cannot generalize well under real-world scenarios. Bridging the synthetic-real data gap remains a reoccurring challenge; see the supplementary material for details.}

Recently, OPD~\cite{jiang2022opd} and its follow-up work OPDMulti~\cite{sun2024opdm}, provide two 2D image datasets of real-world articulated objects: OPDReal and OPDMulti. In OPDReal, images are obtained from frames of RGB-D scans of indoor scenes containing a single object. OPDMulti, on the other hand, captures multiple objects. Both datasets come with 2D segmentation labels on all \emph{openable} parts along with their motion parameters. However, due to the nature of annotation process, the 2D part segmentation masks obtained via 3D-to-2D projection do not fully cover all openable parts in the image. Also, in OPDReal, objects are scanned from within a limited distance range. Practical scenarios and use cases are likely going to have large camera pose and distance variations. OPDMulti, on the other hand, although incorporates such viewpoint variations, a large portion of this dataset contains frames without any articulated objects \cite{sun2024opdm}, which directly affects model training on OPDMulti.

To overcome these limitations, we contribute a 2D image dataset of \xeable objects present in the real world (furniture stores, offices, homes), captured using iPhone 12, 12 Pro and 14. We then use our \emph{coarse-to-fine} AL framework (Figure~\ref{fig:al-pipeline} and Section \ref{sec:method}) to learn generalized 2D segmentations for \xeable object parts.

% Talk about articulation datasets in 3D, including on humans/animals. Then, talk about their 2D counterpart, if there exists any. Here you will need to write about OPDSynth. Then, you talk about OPDReal. Say what issues exist with OPDReal and how our dataset is different from it.
\vspace{-10 pt}
\paragraph{\textbf{Part segmentation in images.}}
Early approaches \cite{wang2015joint, wang2015semantic, xia2017joint} to 2D semantic part segmentation developed probabilistic models on human and animal images. While not addressing the 2D semantic part segmentation problem as such, \cite{huang2020arch, mehta2017vnect, ballan2012motion, kanazawa2018end, mueller2018ganerated} tackled the problem of estimating 3D articulations from human images, which requires an understanding of articulated regions in the input image. 

% Recently, with the availability of 3D part datasets \cite{Mo_2019_CVPR, Xiang_2020_SAPIEN}, there have been works that estimate 3D articulations from articulation images \cite{abbatematteo2019learning, li2020category, zhang2021strobenet, patil2023rosi}. However, most of them learn a latent space of 3D articulation parameters and do not provide 2D segmentation masks for openable parts.
% Our work aims at segmenting \emph{openable} parts of a 3D object from image input. To our knowledge, OPD \cite{jiang2022opd} is the only work that can segment such object parts given an input image, and is built on the Mask RCNN architecture \cite{he2017mask}. In our work, we employ the Mask2Former \cite{cheng2022masked} architecture with task-specific modifications as described in Section \ref{sec:method}.
Recently, the development of large visual models, such as SAM \cite{kirillov2023segany}, has addressed classical 2D vision tasks, such as object segmentation, surpassing all existing models. Such large pre-trained models can be directly employed for \emph{zero-shot} segmentation on new datasets. Follow-up works \cite{zou2023segment, Grounded-SAM_Contributors_Grounded-Segment-Anything_2023} to SAM aim at multi-modal learning by generalizing to natural language prompts. For the task of \xeable part segmentation in real scene images, we observe an unsatisfactory performance using such models. This is expected since they were never trained on any \xeable parts datasets, and therefore, lack an understanding of articulated objects. To our knowledge, OPDMulti~\cite{sun2024opdm} is the SOTA model that can segment \xeable parts in an input image, and is built on the Mask2Former architecture~\cite{cheng2022masked}.
In our work, we use a transformer architecture in a \emph{coarse-to-fine} manner to obtain \xeable part segmentation (Section \ref{sec:method}).

% with the development of LLMs, numerous robust generic models have begun to dominate classic 2D visioin tasks. SAM~\cite{kirillov2023segany} spearheads this trend by demonstrating exceptional capabilities in segmenting unfamiliar objects and images without additional training. There have been works that combine SAM with multi-modal prompts to generalize on open-set tasks~\cite{zou2023segment, Grounded-SAM_Contributors_Grounded-Segment-Anything_2023}. However, without sufficient training data and functionality understanding of articulated objects, these models fail to encode the part information and segment parts in articulated objects correctly. In our work, we aim to segment \xeable parts of multiple articulated objects in images. To our knowledge, OPDFormer~\cite{sun2024opdm} is the state-of-the-art that can segment such parts given an input image, and it built on the Mask2Former architecture~\cite{cheng2022masked}. In our work, we use the hierarchical nature of the problem, leveraging a \emph{coarse-to-fine} Transformer-based segmentation network (Section \ref{sec:method}).} 
% }

\vspace{-10 pt}
\paragraph{\textbf{Active learning for image segmentation.}}
Active learning (AL) is a well-known technique for improving model performance with limited labeled data. \agp{This, in turn, allows the expansion of labeled datasets for downstream tasks.} Prior works \cite{seneractive, Sinha_2019_ICCV, casanovareinforced, Xie_2020_ACCV, Shin_2021_ICCV} have presented different AL frameworks to acquire labels with minimum cost for 2D segmentation tasks. There exist AL algorithms for such tasks \cite{ning2021multi, wu2022d} that are specifically designed to reduce the domain gap by aligning two data distributions. We cannot borrow such methods to reduce the domain gap between synthetic and real scene images of \xeable objects because of the large feature differences: \chk{our synthetic images contain a single object without background or texture, and most objects have an empty interior}.
%Most active learning based algorithms for segmentation tasks denote uncertainty-based active domain adaptation to acquire labels nearby the decision boundary\cite{ning2021multi, wu2022d}, which targets the alignment of two domains. However, the synthetic data of articulated object usually does not contain any meaningful background, which is difficult to align features in the real domain.

More recently, \cite{tang2022active, xie2022towards} employed AL to refine initial 2D segmentation masks through key point or region selection, requiring little human guidance. \chk{These works focus on minimize labeling efforts over the prediction results, whose supervisions effectively correct prediction errors in object masks. However, due} to potentially multiple \xeable parts, such point/region selection is ambiguous for articulated objects. \chk{In our supplementary material, we show that point-based supervision cannot create accurate annotation for \xeable part well.} As such, we design an AL framework \chk{based on our two-stage network} that reduces manual effort by focusing on: (a) using an improved part segmentation model \chk{for generating better samples} (Section~\ref{subsec:network}), and (b) employing a \emph{coarse-to-fine} strategy \chk{to optimize the AL working flow} (Section \ref{subsec:almethod}).  %In our task, the selection of point and region are difficult within a same object, and in most cases, the initial prediction mask has very low quality. Instead of reducing human efforts in annotation, our method focus on getting a better segmentation sampling in AL, with the selection of region-of-interests (ROIs) from pose aware masked attention decoder. }

% \cite{sinha2019variational} introduces a mechanism learning representative sampling in an adversarial manner between unlabeled and labeled data. \cite{shin2021all, cai2021revisiting} involve pixel-level guidance to assist annotations. 

}

%% file: sec/3_problem_statement.tex
\section{Problem Statement}
\label{sec:ps}

\begin{figure}[tb]
\centering
  \includegraphics[width=\textwidth]{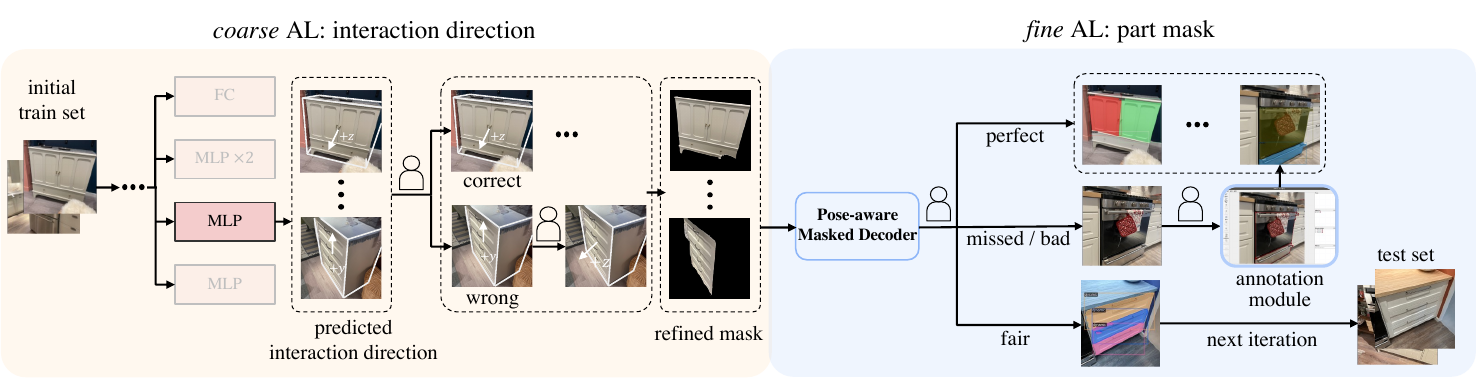}
  % \caption{Our \emph{coarse-to-fine} AL for part segmentation workflow. We first perform \emph{coarse} AL on the interact direction predicted in the \emph{Coarse} stage. With ground truth interact directions, \emph{Fine} stage makes the \xeable part mask predictions. In \emph{fine} AL, we leverage an iteratively training method, which includes human verification and annotation of the part mask, to obtain }
  \caption{Our coarse-to-fine Active Learning (AL) training pipeline. The \emph{coarse} AL applys on interaction directions and retains high-quality predictions while manually rectifying the rest. These rectified predictions form a constructive prior for refined mask prediction. Subsequently, the \emph{fine} AL stage utilizes these refined masks, employing an iterative training method with continuous human intervention for accurate part mask annotation.}
  \label{fig:al-pipeline}
  \vspace{-20 pt}
\end{figure}

% \agp{
% AGP: An interconnection of static and moveable parts compose what are typically referred to as articulated objects. Let $D$ be a real-world image dataset of articulated objects. Given an RGB image $I\in D$ containing one or more articulated objects $\{o_{j}\}$ as input, our goal is to output a set of 2D bounding boxes, $\{b_i\}$, segmentation masks, $\{m_{i}\}$, and semantic labels, $l_i \in \{\text{door}, \text{drawer}\}$ corresponding to all openable parts, $\{p_i\}$, for each object $o\in \{o_{j}\}.$
% An interconnection of static and moveable parts compose what are known as articulated objects
% }
\agp{
Given a set of images $D$ captured from the real-world scene, our input is a single RGB image $I \in D$ containing one or more articulated objects $o_{i}$ from one or more categories $c_{i} \in \{\text{cabinet}, \text{dishwasher}, \text{fridge}, \text{microwave}, \text{oven}, \text{washer}\}$. We assume that each object $o_{i}$ has more than one \xeable parts $P = \{p_1, \dots, p_k\}$ according to its functionality. Our first goal is to predict the 2D bounding box $b_i$, the segmentation mask $m_i$ represented by a 2D polygon and the semantic label $l_i \in \{\text{door}, \text{drawer}\}$ for each \xeable part. 
Extending the above goal, we also aim to build a labeled image dataset that provides accurate 2D segmentation masks and labels for all $p_{i}$'s, for all $I \in D$.
}

%% file: sec/4_method.tex
\section{Method}
\label{sec:method}

\agp{
To address the above problem, we propose an active learning setup that consists of a transformer-based learning framework coupled with a human-in-the-loop feedback process. To this end, we present an end-to-end pose-aware masked-attention network (Fig \ref{fig:method}) that works in a \emph{coarse-to-fine} manner for part segmentation and label prediction. By making use of \emph{coarse} and \emph{fine} features from the network, segmentation masks are further refined by humans in the AL setup (Fig \ref{fig:al-pipeline}), resulting in precise \xeable part masks, \rzz{while minimizing human efforts spent on manual segmentation.}
}
\vspace{-10pt}
\subsection{Pose-aware masked-attention network}
\label{subsec:network}

\agp{
% An overview of our network architecture is shown in Figure \ref{fig:method}. It consists of two stages -- \emph{coarse} and \emph{fine}.
%
% The \emph{coarse} stage, as the name suggests, outputs a coarse segmentation mask (referred to as the ``refined mask" in Figure \ref{fig:method}) on the object's surface that contains a \xeable part. And this surface supports the part's motion, which defines the so-called interaction direction. The \emph{fine} stage leverages this ``refined mask" to output part masks, bounding boxes, and semantic labels, for all \xeable parts in the input image. We explain these two stages below.
}
\rqw{Fig~\ref{fig:method} provides a comprehensive depiction of our network architecture, encompassing two distinct stages. In the \emph{coarse} stage, the network processes a single RGB image and computes a refined mask based on outputs from multiple heads, which accurately pinpoints the region containing \xeable parts. This stage filters out noise predictions on background and extraneous portions of the object. Subsequently, \emph{fine} stage takes the refined mask and image features to generate part masks, bounding boxes, and semantic labels \chk{for all \xeable parts of all articulated objects in the images}.
\vspace{-10pt}
\paragraph{\textbf{Coarse \emph{stage}.}} There are three steps in \emph{coarse} stage. First, the input image is passed through a backbone object detector network \chk{based on MaskRCNN\cite{he2017mask}}, producing multi-scale feature maps $f$ and 2D object bounding boxes $b^{o}$. A pixel decoder~\cite{zhu2020deformable} upsamples $f$ for subsequent processing in the \emph{fine} stage. Second, we use a modified version of the multi-head attention-based encoder and decoder~\cite{zhu2020deformable} to process $f$. Inspired by \cite{jantos2022poet}, we replace the original object query embedding module in \cite{zhu2020deformable} by our new object query embedding with normalized centre coordinates $(c_x, c_y)$, width and height $(w,h)$ from the detected 2D bounding box, enabling the decoder to generate new object query embeddings containing both local global information and estimate 6DoF pose from the 2D bounding box. Third, the decoded queries are passed into multiple \chk{task-specific} MLP heads for (a) object class prediction, (b) 6DoF object pose estimation, (c) object interaction direction prediction and (d) object mask prediction. 

}

\agp{
% \newline
% Object class probabilities are obtained using a fully connected layer followed by a sigmoid activation. 
\rqw{We obtain the object class with \chk{a fully connected network with 3 layers followed by a softmax activation.}} For 6DoF pose estimation, \rqw{we use} two identical MLP heads \chk{with 3 linear layers with ReLU activation} and different output dimensions -- one for estimating camera translation $\tilde{t}=\left(\tilde{t}_{x}, \tilde{t}_{y}, \tilde{t}_{z}\right)$, and the other for estimating the camera rotation matrix $\tilde{R} \in S O(3)$ as described in \cite{zhou2019continuity}. 

The MLP head for interaction direction prediction outputs a set of \chk{6 }possible interaction directions $d \in \{\pm x, \pm y, \pm z\}$ corresponding to the 6DoF coordinates. %\chk{(``6-DoF coordinates"??)}
Using $b^{o}$ and the estimated 6DoF object pose, we can obtain the corresponding 3D \emph{oriented} bounding box $B^{o}$, which tightly fits the $b^{o}$. From among the eight vertices in $B^{o}$, we select vertices of the face along the interaction direction as the representative 2D box for the interaction surface, and use it to crop the input image. \rqw{This cropped image is further multiplied with the object 2D binary mask to filter out background pixels, obtaining the refined binary object mask $m_r$, which guides the subsequent \emph{fine} stage to focus exclusively on the relevant features of the articulated object.}
% This image crop is then multiplied with the 2D binary mask of the object (another MLP head is used to output the 2D binary mask of $o_{i}$) to filter out background pixels. The result is what we refer to as the ``refined mask", $m_{r}$. This completes the \emph{coarse} stage.
}
%
% The interact direction prediction head outputs a set of possible interact directions $d \in \{\pm x, \pm y, \pm z\}$ corresponding to the 6DoF coordinates. Using $b^{o}$ and the estimated 6DoF object pose, we can obtain the corresponding 3D \emph{oriented} bounding box $B^{o}$\rqw{, which tightly fit the $b^{o}$}. From among the eight vertices in $B^{o}$, we select vertices of the face along the interact directions as the representative 2D box for object interact surface, and use it to crop the input image. This cropped image is further multiplied with the 2D binary mask to filter out background pixels, obtaining the refined binary object mask $m_r$, which guides the subsequent \emph{fine} stage to focus exclusively on the relevant features of the articulated object.
\vspace{-10pt}
\paragraph{\textbf{Fine \emph{stage}.}} 
\agp{
There is just one component to this stage, which is the masked-attention decoder from Mask\emph{2}Former \cite{cheng2022masked} (see Figure \ref{fig:method}). It is made up of a cascade of three identical layers, $L_i$'s. $L_{1}$ takes as input image features $f_{pd}$ and the refined mask, $m_r$, and outputs a binary mask which is fed to the next layer. Eventually, the binary mask at the output of $L_{3}$ is multiplied with $f_{pd}$ resulting in \xeable part segmentation in the RGB space. We call this our \emph{pose-aware masked-attention decoder}.
}
\vspace{-10 pt}
\paragraph{\textbf{Loss functions.}}\agp{We formulate the training loss as below
\begin{equation}
    L = L_{class} + L_{dir} + L_{om} + L_{pos} + L_{fine}
\end{equation}
where $L_{class}$ is the binary-cross entropy for object class prediction, $L_{dir}$ is the cross-entropy loss for interaction direction prediction, $L_{om}$ is the binary mask loss for object mask prediction. We define the loss for pose estimation as $L_{pos} = \lambda_t L_t+\lambda_{rot} L_{rot}$, where $L_t$ is the L2-loss of the translation head and $L_{rot}$ is the geodesic loss~\cite{mahendran20173d} of the rotation head. We set $\lambda_t$ and $\lambda_{rot}$ to 2 and 1 respectively. We use a pixel-wise cross-entropy loss for the \emph{fine} stage.
\newline
When pre-training, we jointly train our two-stage network in an end-to-end fashion (see Section~\ref{sec:results}). During fine-tuning on real images with part annotations, we fix MLP weights since ground truth poses and object masks are not available.
}
\subsection{Coarse-to-fine active learning strategy}
\label{subsec:almethod}
\agp{
Our active learning setup, consisting of human-in-the-loop feedback, unfolds in a coarse-to-fine manner (see Figure~\ref{fig:al-pipeline}). We independently run AL workflow on outputs of both \emph{coarse} and \emph{fine} stages from Section \ref{subsec:network}. 

% When employing the AL workflow on the outputs of the \emph{coarse} stage, only the \emph{test set} samples are considered. 
\rqw{In \emph{coarse} AL part, \emph{Coarse} stage generates predictions for the \emph{test set}. In our experiment setup, the \emph{test set} is the \emph{enhancement set}. During this phase, users validate interaction direction predictions and rectify inaccuracies. With ground-truth interaction directions established, refined masks $m_{r}$ are computed and input into the \emph{fine} stage.}
% Here, users validate interaction direction predictions and rectify inaccuracies for every input in the test set, establishing the ground truth. With the now-available ground truth interaction directions, refined masks, $m_{r}$'s, are computed and input into the \emph{fine} stage.

% Our active learning workflow, consisting of human-in-the-loop feedback process on interact direction and part mask, unfolds in a coarse-to-fine manner, as depicted in Figure~\ref{fig:al-pipeline}. We independently learn both \emph{coarse} and \emph{fine} features across two parts. In \emph{coarse} AL part, \emph{Coarse} stage generates predictions for the \emph{test set}. During this phase, users validate interact direction predictions and rectify inaccuracies. With ground-truth interact directions established, refined masks are computed and input into the \emph{fine} stage.

% Employing the AL workflow on the \emph{fine} stage means subjecting the final part segmentation masks and label predictions to user evaluation for one of the following three categories: perfect/missed/fair. 
\rqw{In \emph{fine} AL part, part segmentation mask and label outcome from the \emph{Fine} stage are subject to user evaluation, categorized as perfect, missed, or fair. Specifically: i)} A perfect prediction implies coverage of all \xeable parts in the final segmentation masks, without any gaps, as well as accurate class labels for each segmented part; ii) A missed prediction effectively refers to a null segmentation mask, and/or erroneous class labels; iii) A fair prediction denotes an output segmentation mask that may exhibit imperfections such as gaps or rough edges, and/or may have inaccuracies in some part class labels. \rqw{We provide extensive examples of these scenarios in our supplements.}
%See the supplementary material for details on these. 
During the \rqw{AL} process, perfect predictions are directly incorporated into the next-iteration training set. For all wrong predictions, we employ the labelme~\cite{labelme2016} annotation interface to manually annotate the part mask polygons, and include such images in the next-iteration training set. Fair predictions, on the other hand, remain in the \emph{test set} for re-evaluation.
% In \emph{fine} AL part, part segmentation mask and label outcome from the \emph{Fine} stage are subject to user evaluation, categorized as perfect, missed, or fair. Specifically: i) A perfect prediction implies comprehensive coverage of all \xeable parts by the output segmentation mask without any gaps, coupled with accurate class labels for each part; ii) A wrong prediction is characterized by a total omission of target parts in the output segmentation mask, and/or erroneous class labels; iii) A fair prediction denotes an output segmentation mask that may exhibit imperfections such as gaps or rough edges, and/or may have inaccuracies in some part class labels. We provide extensive examples of these scenarios in our supplementary materials. Throughout our iterative active training process, perfect predictions are directly incorporated into the subsequent training set. For all wrong predictions, we employ the labelme~\cite{yi2016scalable} annotation interface to delineate ground-truth part mask polygons, subsequently including these images in the training set of future iterations. Fair predictions, on the other hand, remain in the \emph{test set} for re-evaluation.

The AL workflow on the \emph{fine} stage continues iteratively until all images within the \emph{test set} transition to the training set, becoming well-labeled and eventually leaving the test set vacant. \rqw{Benefiting from the verified ground-truth interaction direction established in the \emph{coarse} AL part, the \emph{Fine} stage hones in on features of the target surface, omitting noisy object parts. This streamlined focus notably expedites the annotation process. }Further insights into the human verification and annotation procedures will be provided in our supplementary materials.
}

%% file: sec/5_datasets.tex
\section{Datasets and Metrics}
\label{sec:datasets}
\input{tables/data_dist}

\agp{
\paragraph{Datasets.} 
We use three real image datasets in our experiments: (1) OPDReal~\cite{jiang2022opd}, (2) OPDMulti~\cite{sun2024opdm}, and (3) our dataset. 
Our dataset images are obtained from the real world by taking photographs of articulated objects in indoor scenes from furniture stores, offices, and homes, captured using iPhone 12, 12Pro and 14. Images are captured with varying camera poses and distances
from the objects, and an image can contain more than one object, with multiple \xeable parts per object. Differences to OPD and OPDMulti datasets are explained in Section \ref{sec:rw}.

% Different from OPDReal and OPDMulti selecting frames from video data, our dataset comprises images captured through direct photography of indoor scenes in real world. These photographs are rich in detail and resolution, featuring multiple articulated objects with multiple \xeable parts captured from various viewpoints and distances. 

We consider six object categories: Storage, Fridge, Dishwasher, Microwave, Washer, and Oven. A comparison of dataset statistics is presented in Table~\ref{tab:data_dist}. OPDReal comprises of $\sim$30K images, with each image depicting a single articulated object. OPDMulti contains $\sim$64K images. Among these, only 19K images are considered ``valid", containing at least one articulated object. \rqw{Our dataset has a total of 2,550 images, with each image showcasing objects from several categories. We organize our dataset according to the primary object depicted in each image. Our dataset stands out by offering the highest diversity of objects and parts among the compared datasets, including 333 different articulated objects and 1,161 distinct parts. }
% Our dataset has a total of 2,550 images and exhibits larger diversity in terms of constituent objects and their \xeable parts. In total, our dataset consists of 333 different articulated objects and 1,161 distinct parts. 

In terms of the \xeable part annotation, both OPDReal and OPDMulti generate annotations on a 3D mesh reconstructed from the RGB-D scans, and project these 3D annotations back to the 2D image space to get 2D part masks. This process is prone to reconstruction and projection errors. We, on the other hand, create annotations on the captured images directly using our \emph{coarse-to-fine} active learning framework. See the supplementary material for annotation quality comparisons. %We provide a detailed comparison of annotation quality across datasets in the supplementary material. 

Table~\ref{tab:data_dist} shows that the majority (91.67\%) of data samples in OPDReal belong to the Storage category, with the rest distributed among the remaining categories. In contrast, our dataset offers a more uniform data distribution across all six categories. \vspace{-10pt}
\paragraph{Metrics.}
To evaluate model performance and AL efficiency, we use the following:
\begin{itemize}
    \item \textbf{Mean Average Precision (mAP)}: We report mAP@IoU=0.5 for correctly predicting the part label and 2D mask segmentation with IoU $\geq$ 0.5. This metric, which is applied to 2D mask segmentation, is more precise for evaluating segmentation quality than BBox mAP used by OPD~\cite{jiang2022opd}, which only assesses boundary accuracy and overlooks finer details such as mask edges and internal holes.

    \rzz{The {\em ground-truth\/} (GT) segmentations over an image dataset to measure mAP are obtained by applying AL over the dataset with full validation by humans.}
    \item \textbf{AL iterations}: We report the number of iterations required during active learning. This metric represents the efficiency of the overall AL pipeline. 
    \item \textbf{Annotated images}: We report the numbers of images and corresponding parts required for manual annotation for each iteration during AL. This metric helps us evaluate the efficiency of the AL sampling process. 
    \item \textbf{Total lab time}: We report the total lab time required for labeling a dataset. For methods which employ AL, it includes time spent on compulsory sampling after each iteration and manual annotation in each iteration. For methods without AL, it calculates the time spent on manual annotation of all failed predictions. This metric provides an overview of human effort required for all methods. See Section 3.4 in the Supplementary Materials for details of human efforts in our AL process.
    \vspace{-10pt}
\end{itemize}
}
% To evaluate the AL setup, we measure manual annotation \emph{and} verification times. For methods without AL, we record the time for labeling only the \emph{test set}, which is the manual annotation time on network segmentation predictions.

%% file: tables/data_dist.tex
\begin{table}[t]  % Use table* to span the text width
    \centering
   \caption{Dataset statistics across six articulated object categories for OPDReal, OPDMulti and our datasets. Microwave and Oven categories are merged due to their co-occurrence in real scenes. Compared to OPDReal, \agp{our dataset is relatively more balanced in terms of sample distribution of different categories, allowing segmentation models to generalize better}. OPDMulti does not provide category-wise information, and only 19K out of 64K total images are valid with target and annotation. Parts/img shows the average parts annotated for each image. Our dataset exhibits the most object and part diversity among the three datasets.  
}
\vspace{-10pt}
    % \begin{tabularx}{0.85\textwidth}{X*{6}{>{\centering\arraybackslash}X}}
    %     \toprule
    %     & \multicolumn{6}{c}{Category} \\ \cmidrule{3-7}
    %     & & Storage & Fridge & Dishwasher & Micro.\&Oven & Washer \\
    %     \multirow{3}{*}{OPDReal\cite{jiang2022opd}} & Objects & 231 & 12 & 3 & 12 & 3 \\
    %     & Images & 27,394 & 1,321 & 186 & 823 & 159 \\
    %     & image \% & 91.67\% & 3.93\% & 0.62\% & 2.75\% & 0.53\% \\
    %     \cmidrule{2-7}
    %     \multirow{3}{*}{Ours} & Objects & 176 & 51 & 31 & 62 & 13 \\
    %     & Images & 925 & 370 & 315 & 775 & 175 \\
    %     & image \% & 36.27\% & 14.51\% & 12.35\% & 30.39\% & 6.8\% \\
    %     \bottomrule
    % \end{tabularx}

\resizebox{\columnwidth}{!}{
\begin{tabularx}{\textwidth}{@{}ccccccccc@{}}
\toprule
 & \multicolumn{1}{c}{} & \multicolumn{5}{c}{Category} & \multicolumn{1}{c}{Total} & \multicolumn{1}{c}{Parts/img} \\ \cmidrule{3-9} 
 &  & Storage & Fridge & Dishwasher & Mic.\&Oven & Washer &  & \multicolumn{1}{l}{} \\
\multirow{4}{*}{OPDReal\cite{jiang2022opd}} & Objects & 231 & 12 & 3 & 12 & 3 & 284 & \multirow{4}{*}{2.22}\\
 & Images & 27,394 & 1,321 & 186 & 823 & 159 & 30K \\
 & image \% & 91.67\% & 3.93\% & 0.62\% & 2.75\% & 0.53\% & 100\% \\
 & Parts & 787 & 27 & 3 & 13 & 3 & 875 \\ \cmidrule{2-9} 
\multirow{3}{*}{OPDMulti\cite{sun2024opdm}} & Objects & - & - & - & - & - & 217 & \multirow{3}{*}{1.71}\\
 & Images & - & - & - & - & - & 19K/64K \\
 & Parts & - & - & - & - & - & 688 \\ \cmidrule{2-9} 
\multirow{4}{*}{Ours} & Objects & 176 & 51 & 31 & 62 & 13 & 333 & \multirow{4}{*}{4.33}\\
 & Images & 925 & 370 & 315 & 775 & 175 & 2550 \\
 & image \% & 36.27\% & 14.51\% & 12.35\% & 30.39\% & 6.8\% & 100\% \\
 & Parts & 896 & 159 & 31 & 62 & 13 & 1161 \\ \bottomrule
\end{tabularx}
}
\label{tab:data_dist}
\vspace{-20pt}
\end{table}

%% file: sec/6_results.tex
\section{Experiments}
\label{sec:results}
\rqw{
\agp{We start our experiments by rendering synthetic images from the PartNet-Mobility dataset~\cite{Xiang_2020_SAPIEN} with diverse articulation states, enabling us to obtain sufficient annotations for training 2D segmentation networks and support transfer learning applications.} The synthetic dataset contains $\sim$32K images, evenly distributed across categories, and randomly partitioned into training  (90\%) and test sets (10\%). 

We implement our network in PyTorch on two NVIDIA Titan RTX GPUs. \agp{All images are resized to 256$\times$256 for training.} For pre-training on PartNet-Mobility, we use the Adam optimizer with an initial learning rate (\emph{lr}) of 2.5$e$-4, reducing it by %a factor of
$\gamma=0.1$ at 1K and 1.5K epochs separately over a total of 2K epochs. When fine-tuning on real images, we use the same \emph{lr} and $\gamma$ at 3.5K and 4K epochs over a total of 4.5K epochs.

\subsection{Competing methods}
We compare our active \textit{coarse-to-fine} part segmentation model with three 2D segmentation methods and also analyze two variants of our proposed approach. 
\begin{itemize}
    \item \textbf{Grounded-SAM}~\cite{Grounded-SAM_Contributors_Grounded-Segment-Anything_2023}, which combines Grounding-DINO~\cite{liu2023grounding} and Segment Anything~\cite{kirillov2023segany}, \agp{is a \rzz{foundational vision-language} model that can be used for zero-shot 2D object detection and segmentation, and supports text prompts.} \chk{In our experiments, we set the text prompt as [door, drawer] for segmentation results. } %as a powerful large model used for strong zero-shot object detection and segmentation in images given text captions.
    \item \textbf{OPD-C}~\cite{jiang2022opd}, which is the first work for detecting openable parts in images based on MaskRCNN~\cite{he2017mask}. This is the base variant without camera pose for training. 
    \item \textbf{OPDFormer-C}~\cite{sun2024opdm}, a follow-up of OPD-C based on Mask2Former~\cite{cheng2022masked}, is the SOTA for \agp{openable} part detection of multiple articulated objects in images.
% \end{itemize}
% \vspace{-10pt}
% And two variants of our approach,
% \begin{itemize}
    \item \textbf{$\text{Ours}_{w/o AL}$} \agp{is a variant that does not use human feedback -- it infers part segmentation results based only on the transformer-based model.} %\emph{test set} using our transformer-based model.
    \item \textbf{$\text{Ours}_{f-AL}$} \agp{is variant of our approach that uses only the \emph{fine} stage of the AL framework. That is, verification and annotation of just the part masks is done.}
    % \item \textbf{$\text{Ours}_{GAL}$} \agp{is an AL variant of our approach \emph{without} using the \emph{coarse-to-fine} AL strategy -- verification and annotations of just the part masks is done (\emph{fine} stage only).}
    %is an AL variant of our approach without using the \emph{coarse-to-fine} AL strategy; it is trained using the \emph{fine} AL with verification and annotation of part mask only.
    \vspace{-10pt}
\end{itemize}

\input{tables/mAP_ours}

\subsection{Evaluation on Our Dataset}

\rzz{We perform two key evaluations on our dataset, one for segmentation accuracy and one for
annotation efficiency, while comparing to SOTA alternatives, with or without AL.}

\rzz{
%\paragraph{Dataset split.}
%
We work on 2,550 images with a split of 50/500/2000 into \textit{train/enhancement/test} sets. The train set has been fully annotated manually and
it is used by all methods, except for Grounded-SAM, for fine-tuning. The
\emph{enhancement set}, initially unlabeled, is employed by AL models to progressively improve the learning. The \emph{test set} of 2,000 images is unseen by all methods, including AL, when evaluating segmentation accuracy. When assessing annotation efforts, we apply the methods on both the 500-image set and the 2,000-image set to examine how the efficiency achieved by our AL model scales.
%This demonstrates the initial scarcity of annotated data. We first evaluate the performance of all competing methods on our . Note that  Methods employing AL utilize our \emph{enhancement set} for further learning. In addition, we evaluate the annotation efficiency of all methods, with and without AL, on the \emph{enhancement set} and the \emph{test set} respectively. 
}

\vspace{-5pt}

\paragraph{\rzz{Segmentation accuracy} on \emph{test set}.} Table~\ref{tab:map_ours} compares four non-AL and two AL methods. Among four non-AL methods (columns 1-4), Grounded-SAM is without fine-tuning and \agp{has the lowest performance}. This demonstrates that current \rzz{generic large foundational models} are still limited in \agp{understanding} object parts without adequate \agp{training on} well-labeled data. 
\rzz{Despite the small (50-image) \emph{train set}, models fine-tuned on it produce significant improvements. Specifically, $\text{Ours}_{w/o AL}$ model surpasses all competing methods with over 75\% segmentation mAP, while OPDFormer-C falls short of 70\%, and OPD-C scores below 50\%. This discrepancy stems from the architectural designs of OPD-C and OPDFormer-C, which were built on vanilla MaskRCNN and Mask2Former for general segmentation tasks but fail to capture the nuances of articulated objects, where movable parts are closely tied to object pose and interaction directions.} In contrast, our network effectively leverages the hierarchical structure of the scene, objects, and parts therein, resulting in \agp{a much better} performance. 

% Conversely, $\text{Ours}_{w/o AL}$ model outperforms competing methods, achieving $>$75\% segmentation mAP. The \agp{architectural} designs of OPD-C and OPDFormer, based respectively on vanilla MaskRCNN and Mask2Former, are tailored for universal segmentation tasks. Therefore, they inadequately address the unique characteristics of articulated objects, whose movable parts are closely tied to object pose and interact\agp{ion} directions. 
%Beyond accuracy metrics, we observe that all non-AL methods require \agp{significant} human effort for labeling the \emph{test set}. \agp{This is mainly due} to imperfect predictions from these models. 

As seen in the last two columns of Table~\ref{tab:map_ours}, \chk{by performing AL on the \emph{enhancement set}}, \rzz{the performance is significantly boosted over non-AL methods,} reaching over 90\% accuracy, with \chk{less than 1.7 hours spent on} manual \rzz{segmentation}. 
Figure~\ref{fig:viz_ours} shows qualitative results of different methods on our \emph{test set}.

The segmentation accuracy of our two AL alternatives is close since they share the identical network architecture. But they differ in AL training strategies, which impacts labeling efficiency. \rzz{On the 500-image \emph{enhancement set}, our \emph{coarse-to-fine} AL strategy leads to a slight improvement (only 4.5\%) on human annotation effort. We show next that on a larger set to perform AL, the improvement becomes more significant.}

\vspace{-5pt}

\input{tables/al_ours}

\paragraph{Annotation efficiency comparison.}
\rzz{Table~\ref{tab:map_ours} shows that with 1.6 hours of manual segmentation to process images with missed predictions, our AL model is able to fully validate the \xeable part segmentations and semantic labels for the 500-image \emph{enhancement set}. 
To obtain the same GT annotations on this set, with respect to an non-AL method such as Grounded-SAM, one must manually correct all images with erroneous or imperfect segmentations.
Specifically, Grounded-SAM could only yield less than 5\% perfectly annotated images, with the rest (479) needing manual processing.

In Table~\ref{tab:al_ours} (top), we report and compare manual efforts, in terms of number of images, parts, and annotation times, required across different methods to obtain GT for the 500-image set.
%
%annotating images in this set. As shown in row 1-4, without AL, at least over 200 images require manual labeling. Specifically, Grounded-SAM and OPD-C have less than 30 perfect predictions, requiring more than 7.5 hours to manually label the remaining images. 
$\text{Ours}_{w/o AL}$ shows the best efficiency among non-AL methods, but it still takes 3.58 hours to annotate 210 images with 762 parts. Rows 5-8 underscore the benefits of AL for  annotation efficiency. By employing AL, in rows 5 \& 6, OPD-C and OPDFormer-C demonstrate marked improvements over their non-AL versions. However, they still require 5.7 and 3.9 hours, respectively. Due to their tendency to generate noisy predictions on irrelevant parts of the object or background, most predictions are categorized as \emph{fair} as described in Section~\ref{subsec:almethod}, leading to more iterations in AL and additional time spent on sampling. In contrast, as shown in rows 7 \& 8, both variants of our AL model complete in 3 iterations, with our \emph{coarse-to-fine} AL methods requiring the least images for labeling and minimum time efforts.

In Table~\ref{tab:al_ours} (bottom), we report all the numbers to obtain GT annotations for a larger set of (2,000) images. What is most notable is that the efficiency gain in manual annotation time by our \emph{coarse-to-fine} AL strategy has improved from less than 5\%, for the smaller image set of 500, to more than 13\% (6.5 hours vs.~7.5 hours). This demonstrates that our \emph{coarse-to-fine} AL approach is particularly beneficial for large-scale annotation tasks, where the time saved on annotation significantly outweighs the extra time spent on sampling. Please check our supplement for detailed AL iterations.
}

%\chk{To further assess the efficiency of AL methods, Table~\ref{tab:al_ours} (bottom) compares time savings between non-AL methods (row 1-4) and our AL methods (row 5 \& 6) for annotating our \emph{test set}, which contains more data. We can observe that the advantages of AL methods are more notable. When evaluating results on our \emph{enhancement set}, our \emph{coarse-to-fine} AL method shows marginal time savings of 0.005 hour over $\text{Ours}_{f_AL}$, as it requires an additional iteration for sampling interaction directions. The benefit of our \emph{coarse-to-fine} AL strategy becomes more pronounced with larger datasets; the time saved is more substantial as the volume of data needing annotation increases. Specifically, for 2000 images, the time saving escalates to 1 hour, highlighting the scalability and efficiency advantages of the method in handling extensive datasets. This demonstrates that our \emph{coarse-to-fine} AL approach is particularly beneficial for large-scale annotation tasks, where the time saved on annotation significantly outweighs the extra time spent on sampling. Please check our supplement for detailed AL iterations.}

\input{tables/ablation}

\vspace{-5pt}

\paragraph{Ablation study.}Table~\ref{tab:ablation} \agp{highlights the need and} contributions of key components of our method on improving prediction accuracy and minimizing human efforts. Columns 2-5 respectively indicate the presence of: \textbf{Mask} object mask head; \textbf{Pose} object pose estimation head; \textbf{Interaction direction} interact\agp{ion} direction prediction head; \textbf{AL} active learning. Row 4 and 5 uses the \emph{fine} AL stage on part mask \agp{alone} due to the absence of pose and interact\agp{ion} direction prediction module. Row 6 and 7 use our \emph{coarse-to-fine} AL strategy. Results in Table~\ref{tab:ablation} clearly justify the \emph{coarse-to-fine} design in our method, \agp{which gives the best performance (see row 7).} %where our method (Row 6) is the most accurate and efficient.

\subsection{Evaluation on OPDReal and OPDMulti}
In addition, we assess the performance of \agp{different} models on the OPDReal and OPDMulti dataset, using \agp{their respective train and test splits.} %training and test data partitions as provided by these datasets. 

\begin{wraptable}{r}{6cm}
\vspace{-25pt}
\caption{Quantitative comparison against competing segmentation methods and our model variant on OPDReal, OPDMulti test set.}\label{tab:map_opd}
\resizebox{0.5\columnwidth}{!}{
\begin{tabular}{@{}ccccc@{}}
\toprule
& \multicolumn{3}{c}{segm mAP ($\uparrow$)}\\
\cmidrule{2-5}
& Grounded-SAM & OPD-C & OPDFormer-C & $\text{Ours}_{w/o AL}$ \\
\cmidrule{2-5}
OPDReal & 16.5 & 44.5 & 46.3 & \textbf{51.6}\\
OPDMulti & 8.0 & 25.6 & 27.6 & \textbf{31.5}\\
\bottomrule
\end{tabular}
}
\vspace{-20pt}
\end{wraptable} 

As shown in Table~\ref{tab:map_opd}, $\text{Ours}_{w/o AL}$ method \agp{outperforms the rest}. 
However, with more than 70\% training data, all methods still fail to achieve $>$55\% accuracy on OPDReal. This limitation primarily stems from \agp{data skewness towards the Storage category in OPDReal, which constitutes more than 90\% of total samples, and results in poor generalization across other object categories.} Detailed category-wise results are \agp{provided} in the supplementary material. 

Performance on OPDMulti is further compromised by an abundance of noisy data in its test set\agp{\cite{sun2024opdm}}. From the qualitative results in Figure~\ref{fig:viz_opdreal}, we observe that some openable parts are cluttered or missed in the GT annotation, while our method accurately segments these parts. This discrepancy also contributes to the low accuracy.
}

\begin{figure}[tb]
    \centering
    \includegraphics[width=\textwidth]{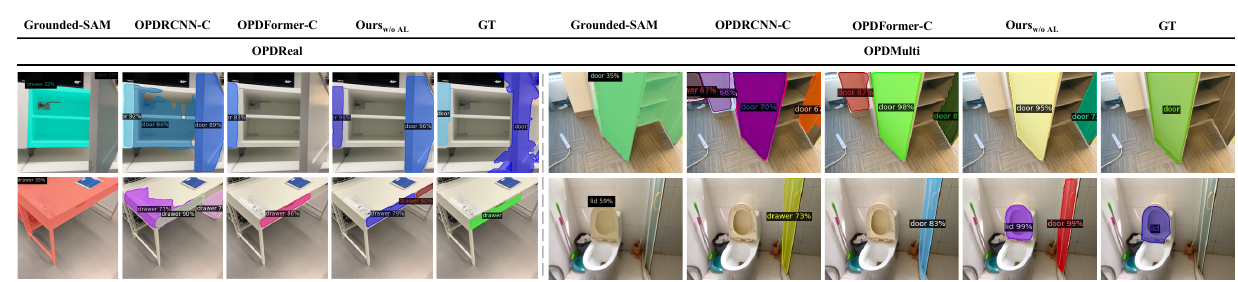}  
    \caption{Qualitative results on OPDReal and OPDMulti test set. $\text{Ours}_{w/o AL}$ outperforms others on noisy GT and multiple objects. \chk{See supplementary materials for more results.}}
    \label{fig:viz_opdreal}
    \vspace{-10pt}
\end{figure}
\begin{figure}[t]
    \centering
    \includegraphics[width=\textwidth]{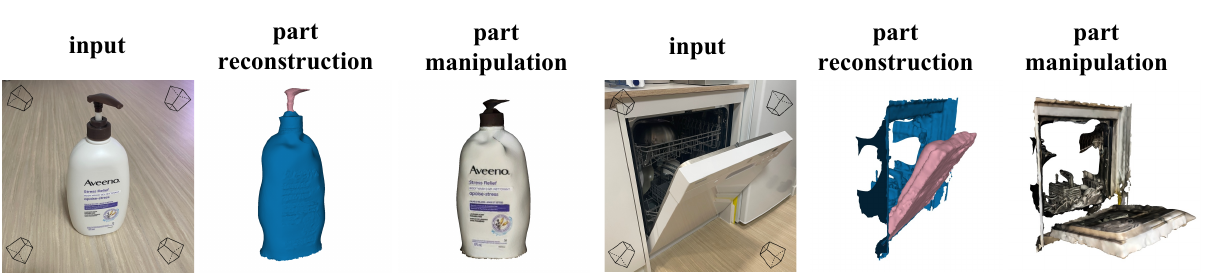}
    \caption{Part-level reconstruction and manipulation of the bottle and dishwasher}
\label{fig:application}
\vspace{-10pt}
\end{figure}

%% file: tables/mAP_ours.tex
%% Grounded-SAM: no finetuning, test on 2K test set directly; OPD-C, M2F, ours w/o AL: train on 50, test on 2k test set; ours w AL: start with 50, iteratively labeling all 500 enhancement set, then test on 2K test set.
\begin{table}[t]\centering
% \resizebox{\columnwidth}{!}{
% \begin{tabular}{@{}cccc@{}}
% \toprule
% & \multicolumn{3}{c}{segm mAP / bbox mAP ($\uparrow$)}\\
% \cmidrule{2-4}
% & OPD-C\cite{jiang2022opd} & M2F\cite{cheng2022masked} & Ours \\
% \cmidrule{2-4}
% OPDReal & 37.380 / 44.663 & 39.380 / 42.153 & \textbf{48.168} / \textbf{54.393}\\
% Ours & 85.702 / 86.094 & 93.245 / 94.052 &  \textbf{96.578} / \textbf{96.756}   \\
% \bottomrule
% \end{tabular}
% }
% \begin{tabular}{cccc@{}}
% \toprule
% Method & segm mAP $\uparrow$ & Imgs / Parts$\downarrow$ & Time (hr)$\downarrow$ \\
% \midrule
% Grounded-SAM\cite{Grounded-SAM_Contributors_Grounded-Segment-Anything_2023} &  22.9 & 1942/6456 & 26.9\\
% OPD-C\cite{jiang2022opd} & 46.4 & 1843 / 6095 & 25.4\\
% Mask2Former\cite{cheng2022masked} & 71.8 & 1072 / 3651 & 15.2\\
% Ours w/o AL & 82.1 & 564 / 1936 & 8.1 \\
% Ours w AL & \textbf{96.7} & \textbf{229 / 758} & \textbf{6.5} \\
% \bottomrule
% \end{tabular}
% }
% \begin{tabular}{ccc@{}}
% \toprule
% Method & bbox mAP $\uparrow$  & segm mAP $\uparrow$ \\
% \midrule
% Grounded-SAM\cite{Grounded-SAM_Contributors_Grounded-Segment-Anything_2023} & 25.5 & 22.9 \\
% OPD-C\cite{jiang2022opd} & 48.6 & 46.4 \\
% Mask2Former\cite{cheng2022masked} & 73.1 & 71.8 \\
% Ours w/o AL & 82.6 & 82.1 \\
% Ours w AL & \textbf{96.7} & \textbf{96.7} \\
% \bottomrule
% \end{tabular}

\caption{\rzz{Comparing segmentation accuracy against competing methods and variants of our method on the \emph{unseen test set} of 2,000 real images. In the table, ``AL'' indicates whether the method uses active learning. All methods take the \textit{train} set as the training data, and are evaluated on the \textit{test set}. Methods in the last two columns perform AL on the \textit{enhancement set}. The ``Time'' row represents the total lab time metric described in Section \ref{sec:datasets}, only for methods using AL.}} 
\vspace{-10pt}
\resizebox{\columnwidth}{!}{
\begin{tabular}{ccccccc@{}}
\toprule
& \multicolumn{6}{c}{Method} \\
\cmidrule{2-7}
& Grounded-SAM\cite{Grounded-SAM_Contributors_Grounded-Segment-Anything_2023} &  OPD-C\cite{jiang2022opd}  & OPDFormer-C\cite{sun2024opdm} & $\text{Ours}_{w/o AL}$ & $\text{Ours}_{f-AL}$ & Ours\\
\midrule
AL & -  & - & - & - & \checkmark & \checkmark \\
segm mAP $\uparrow$ & 23.1 & 45.2 & 68.4 & 77.3 &  91.2 & \textbf{91.3}\\
Time (hr) $\downarrow$ & - & - & - & - & 1.675 & \textbf{1.6} \\
\bottomrule
\end{tabular}
}
\label{tab:map_ours}
\vspace{-10pt}
\end{table}

%% file: tables/al_ours.tex
\begin{table}[t]
\centering
\caption{\rzz{Comparison of manual segmentation efforts required for different methods in annotating segmentation masks for two sets of images of different sizes.1 In the table, ``AL'' indicates whether the method uses active learning for labeling. All methods are trained on the original 50-image \emph{train} set, when annotating the 500-image set, When annotating the 2,000-image set, we add the 500 images with ground-truth segmentations to the train set (50+500=550 images).}}
\vspace{-10pt}
\label{tab:al_ours}
% \resizebox{\columnwidth}{!}{
\centering
      \begin{tabular}{@{}ccccccc@{}}
        \toprule 
        Dataset size & Row ID & Method & AL & Iterations & \#Images / Parts. & Time (hr) $\downarrow$\\
        \midrule 
        \multirow{8}{*}{\shortstack{\rzz{500 images}}}
        & 1 & Grounded-SAM & - & - & 479 / 1,711 & 7.54 \\
        & 2 & OPD-C & - & - & 483 / 1,720 & 7.58 \\
        & 3 & OPDFormer-C & - & - & 324 / 1,248 & 5.62 \\
        & 4 & $\text{Ours}_{w/o AL}$ & - & - & 210 / 762 & 3.58\\
        & 5 & OPD-C & \checkmark & 7 & 261 / 887 & 5.7 \\
        & 6 & OPDFormer-C & \checkmark & 7 & 158 / 560 & 3.9 \\
        & 7 & $\text{Ours}_{f-AL}$ & \checkmark & \textbf{3} & 83 / 239 & 1.675 \\
        & 8 & Ours & \checkmark & \textbf{3} & \textbf{64 / 184} & \textbf{1.6}\\
        \cmidrule{1-7}
       \multirow{6}{*}{\shortstack{\rzz{2,000 images}}}
        & 1 & Grounded-SAM & - & - & 1,888 / 8,119 & 35.5\\
        & 2 & OPD-C & - & - & 1,420 / 6,391 & 28.3\\
        & 3 & OPDFormer-C & - & - & 792 / 3,511 & 16.3\\
        & 4 & $\text{Ours}_{w/o AL}$ & - & - & 625 / 2,815 & 13.4\\
        & 5 & $\text{Ours}_{f-AL}$ & \checkmark & \textbf{4} & 289 / 983 & 7.5\\
        & 6 & Ours & \checkmark & \textbf{4} & \textbf{229 / 158}& \textbf{6.5}\\
        \bottomrule
      \end{tabular}
% }
\vspace{-20pt}
\end{table}

%% file: tables/ablation.tex
\begin{table}[t]\centering
\caption{Ablation study on our key components.}
% \begin{tabular}{@{}cccc@{}}
% \toprule
% Row ID & Object Mask & Pose & segm mAP ($\uparrow$)\\
% \midrule
% 0 & - & - & 93.245  \\
% 1 & \checkmark & - &  94.421  \\
% 2 & - & \checkmark &  96.025  \\
% 3 & \checkmark & \checkmark & \textbf{96.672} \\
% \bottomrule
% \end{tabular}
\vspace{-10pt}
\resizebox{0.8\columnwidth}{!}{
\begin{tabular}{@{}ccccccc@{}}
\toprule
Row ID & Mask & Pose & Interaction direction & AL & segm mAP $\uparrow$ & Time (hr) $\downarrow$\\
\midrule
1 & - & - & - & - & 68.4 & -\\
2 & \checkmark & - & - & - & 74.9 & - \\
3 & \checkmark & \checkmark & \checkmark & - & 77.3 & -\\
4 & - & - & - & \checkmark & 87.3 & 3.9\\
5 & \checkmark & - & - & \checkmark & 89.1 & 3.1\\
6 & - & \checkmark & \checkmark & \checkmark & 90.8 & 2.4\\
7  & \checkmark & \checkmark & \checkmark & \checkmark & \textbf{91.3} & \textbf{1.6}\\
\bottomrule
\end{tabular}
}
\vspace{-15pt}
\label{tab:ablation}
\end{table}

%% file: sec/7_application.tex
\section{Application}
\label{sec:app}

\rqw{
Our work demonstrate practical applications in part based reconstruction and manipulation of articulated objects from images. Given a set of multi-view RGB images of an articulated object, our model predicts precise segmentation masks of \xeable parts in each image. This enables part based 3D reconstruction using masked images for both \xeable parts and the main body of the object. The resulting 3D models of parts allow for easy manipulation of \xeable parts to unseen states in 3D as shown in Figure~\ref{fig:application}.

}

%% file: sec/8_conclusion.tex
\section{Conclusion}
\label{sec:conclusion}
We present the first active segmentation framework for high-accuracy instance segmentation of \xeable parts in real-world RGB images. Our active learning framework, integrating human feedback, iteratively refines predictions in a \emph{coarse-to-fine} manner, and \agp{achieves close-to-error-free performance on the test set}. By leveraging correlations between the scene, objects, and parts, we demonstrate that our method can achieve state-of-the-art performance on challenging scenes with multiple cross-categories objects, and \agp{significantly reduce human efforts for dataset preparation.}

Additionally, we \agp{contribute a high-quality and diverse dataset} of articulated objects in real-world scene, complete with precise \xeable part annotations. We will expand it further to support the vision community for understanding scene from images. We also hope our work catalyzes future motion- or functionality-aware vision tasks.
% \agp{
% We advocate active learning as a general and effective means to obtain high-accuracy instance segmentations. It may be the most
% viable option to achieve close-to-error-free performance on arbitrary test sets. If properly designed, AL can significantly reduce
% human annotation effort for dataset preparation. In this work, we realized both goals for the specific task of instance segmentation
% of interactable parts from real scene images containing articulated objects.

% Our contribution also includes a high-quality and diverse dataset of annotated real photographs, which we will continue to scale up
% to serve the vision community. We would also like to endow the annotated parts with motion parameters. On the technical side, there is
% much room to improve on speeding up the correction of erroneous segmentations during AL. Additional priors beyond object poses
% may also be explored to facilitate dynamic part segmentation. At last, we would like to extend our AL framework to other motion-
% or functionality-aware vision and annotation tasks.
% }

\begin{figure}[t]
    \centering
    \includegraphics[width=\textwidth]{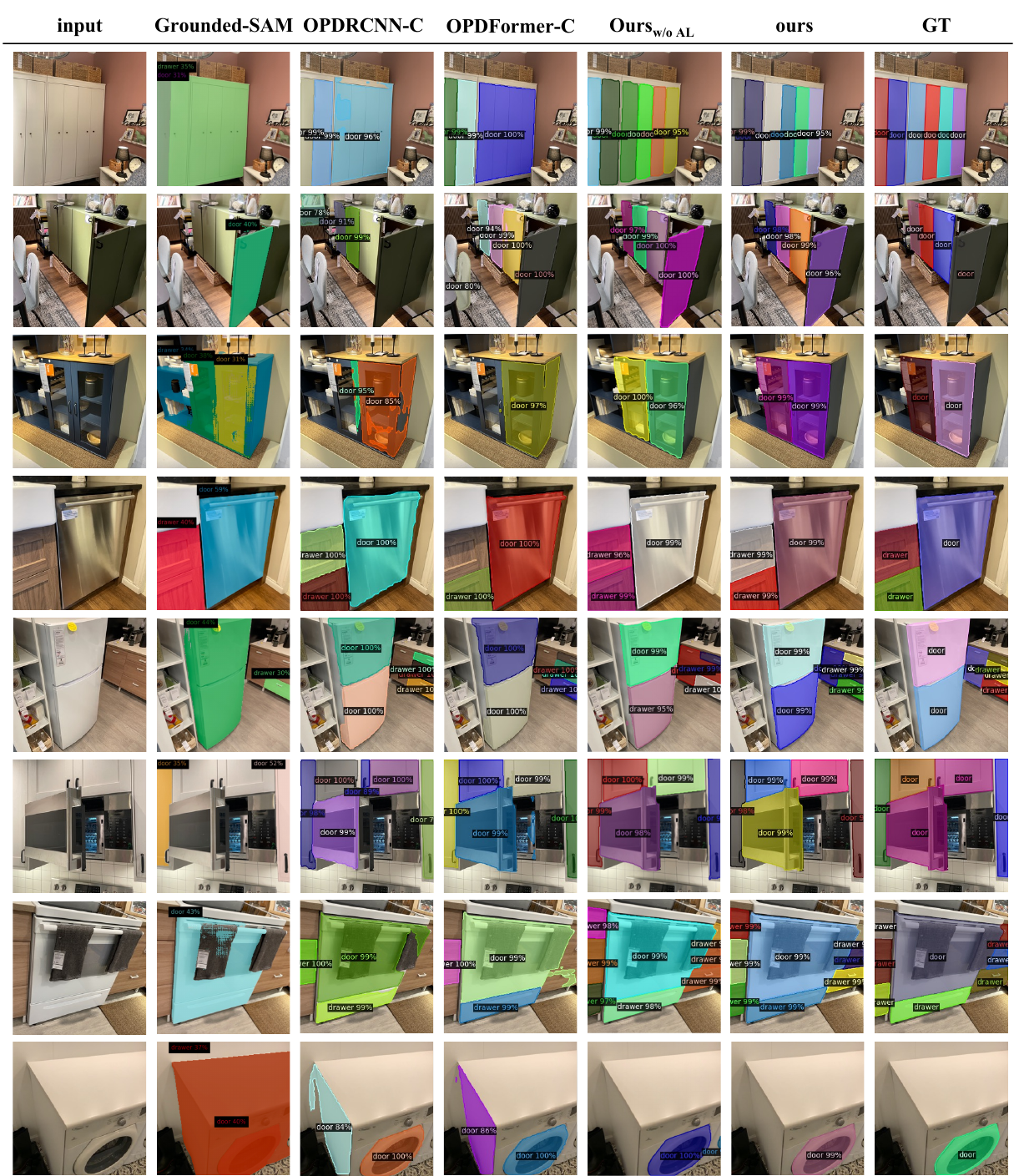}
    % \vspace{-3pt}
    \caption{Qualitative results on \textit{test set} from our dataset. We visualize predictions results on different object categories using 3 competing methods and our final model. Our method outputs better segmentation masks over \xeable parts across multiple objects in the image with clear separation of parts and small parts segmentation (Row 1, 4, 5). Our results also show that the \emph{coarse-to-fine} segmentation framework can effectively reduce segmentation errors from unwanted objects (Row 2) and object side surfaces (Row 2, 3, 6, 8). More results in the supplementary materials. }
    \label{fig:viz_ours}
\end{figure}

%% file: main.bbl
\begin{thebibliography}{10}
\providecommand{\url}[1]{\texttt{#1}}
\providecommand{\urlprefix}{URL }
\providecommand{\doi}[1]{https://doi.org/#1}

\bibitem{trimble_warehouse}
{Trimble Inc. 3D Warehouse}. \url{https://3dwarehouse.sketchup.com/} (2023), accessed: 2023-3-4

\bibitem{AL_survey2014}
Aggarwal, C.C., Kong, X., Gu, Q., Han, J., Yu, P.S.: Active learning: A survey. In: Data Classification: Algorithms and Applications, pp. 571--597 (2014)

\bibitem{ballan2012motion}
Ballan, L., Taneja, A., Gall, J., Van~Gool, L., Pollefeys, M.: Motion capture of hands in action using discriminative salient points. In: Computer Vision--ECCV 2012: 12th European Conference on Computer Vision, Florence, Italy, October 7-13, 2012, Proceedings, Part VI 12. pp. 640--653. Springer (2012)

\bibitem{casanovareinforced}
Casanova, A., Pinheiro, P.O., Rostamzadeh, N., Pal, C.J.: Reinforced active learning for image segmentation. In: International Conference on Learning Representations (2020)

\bibitem{chang2015shapenet}
Chang, A.X., Funkhouser, T., Guibas, L., Hanrahan, P., Huang, Q., Li, Z., Savarese, S., Savva, M., Song, S., Su, H., et~al.: Shapenet: An information-rich 3d model repository. arXiv preprint arXiv:1512.03012  (2015)

\bibitem{cheng2022masked}
Cheng, B., Misra, I., Schwing, A.G., Kirillov, A., Girdhar, R.: Masked-attention mask transformer for universal image segmentation. In: Proceedings of the IEEE/CVF Conference on Computer Vision and Pattern Recognition. pp. 1290--1299 (2022)

\bibitem{dosovitskiy2020image}
Dosovitskiy, A., Beyer, L., Kolesnikov, A., Weissenborn, D., Zhai, X., Unterthiner, T., Dehghani, M., Minderer, M., Heigold, G., Gelly, S., et~al.: An image is worth 16x16 words: Transformers for image recognition at scale. In: ICLR (2021)

\bibitem{Grounded-SAM_Contributors_Grounded-Segment-Anything_2023}
{Grounded-SAM Contributors}: {Grounded-Segment-Anything} (Apr 2023), \url{https://github.com/IDEA-Research/Grounded-Segment-Anything}

\bibitem{he2017mask}
He, K., Gkioxari, G., Doll{\'a}r, P., Girshick, R.: Mask r-cnn. In: Proceedings of the IEEE international conference on computer vision. pp. 2961--2969 (2017)

\bibitem{hu2017learning}
Hu, R., Li, W., Van~Kaick, O., Shamir, A., Zhang, H., Huang, H.: Learning to predict part mobility from a single static snapshot. ACM Transactions on Graphics (TOG)  \textbf{36}(6),  1--13 (2017)

\bibitem{huang2021multibodysync}
Huang, J., Wang, H., Birdal, T., Sung, M., Arrigoni, F., Hu, S.M., Guibas, L.J.: Multibodysync: Multi-body segmentation and motion estimation via 3d scan synchronization. In: Proceedings of the IEEE/CVF Conference on Computer Vision and Pattern Recognition. pp. 7108--7118 (2021)

\bibitem{huang2020arch}
Huang, Z., Xu, Y., Lassner, C., Li, H., Tung, T.: Arch: Animatable reconstruction of clothed humans. In: Proceedings of the IEEE/CVF Conference on Computer Vision and Pattern Recognition. pp. 3093--3102 (2020)

\bibitem{jantos2022poet}
Jantos, T., Hamdad, M., Granig, W., Weiss, S., Steinbrener, J.: {PoET: Pose Estimation Transformer for Single-View, Multi-Object 6D Pose Estimation}. In: 6th Annual Conference on Robot Learning (CoRL 2022) (2022)

\bibitem{jiang2022opd}
Jiang, H., Mao, Y., Savva, M., Chang, A.X.: Opd: Single-view 3d openable part detection. In: Computer Vision--ECCV 2022: 17th European Conference, Tel Aviv, Israel, October 23--27, 2022, Proceedings, Part XXXIX. pp. 410--426. Springer (2022)

\bibitem{kanazawa2018end}
Kanazawa, A., Black, M.J., Jacobs, D.W., Malik, J.: End-to-end recovery of human shape and pose. In: Proceedings of the IEEE conference on computer vision and pattern recognition. pp. 7122--7131 (2018)

\bibitem{kirillov2023segany}
Kirillov, A., Mintun, E., Ravi, N., Mao, H., Rolland, C., Gustafson, L., Xiao, T., Whitehead, S., Berg, A.C., Lo, W.Y., Doll{\'a}r, P., Girshick, R.: Segment anything. arXiv:2304.02643  (2023)

\bibitem{li2020category}
Li, X., Wang, H., Yi, L., Guibas, L.J., Abbott, A.L., Song, S.: Category-level articulated object pose estimation. In: Proceedings of the IEEE/CVF Conference on Computer Vision and Pattern Recognition. pp. 3706--3715 (2020)

\bibitem{liu2023grounding}
Liu, S., Zeng, Z., Ren, T., Li, F., Zhang, H., Yang, J., Li, C., Yang, J., Su, H., Zhu, J., et~al.: Grounding dino: Marrying dino with grounded pre-training for open-set object detection. arXiv preprint arXiv:2303.05499  (2023)

\bibitem{mahendran20173d}
Mahendran, S., Ali, H., Vidal, R.: 3d pose regression using convolutional neural networks. In: Proceedings of the IEEE International Conference on Computer Vision Workshops. pp. 2174--2182 (2017)

\bibitem{mehta2017vnect}
Mehta, D., Sridhar, S., Sotnychenko, O., Rhodin, H., Shafiei, M., Seidel, H.P., Xu, W., Casas, D., Theobalt, C.: Vnect: Real-time 3d human pose estimation with a single rgb camera. Acm transactions on graphics (tog)  \textbf{36}(4),  1--14 (2017)

\bibitem{Mo_2019_CVPR}
Mo, K., Zhu, S., Chang, A.X., Yi, L., Tripathi, S., Guibas, L.J., Su, H.: {PartNet}: A large-scale benchmark for fine-grained and hierarchical part-level {3D} object understanding. In: The IEEE Conference on Computer Vision and Pattern Recognition (CVPR) (June 2019)

\bibitem{mueller2018ganerated}
Mueller, F., Bernard, F., Sotnychenko, O., Mehta, D., Sridhar, S., Casas, D., Theobalt, C.: {GAN}erated hands for real-time 3d hand tracking from monocular rgb. In: Proceedings of the IEEE conference on computer vision and pattern recognition. pp. 49--59 (2018)

\bibitem{ning2021multi}
Ning, M., Lu, D., Wei, D., Bian, C., Yuan, C., Yu, S., Ma, K., Zheng, Y.: Multi-anchor active domain adaptation for semantic segmentation. In: Proceedings of the IEEE/CVF International Conference on Computer Vision. pp. 9112--9122 (2021)

\bibitem{AL_survey2020}
Ren, P., Xiao, Y., Chang, X., Huang, P.Y., Li, Z., Gupta, B.B., Chen, X., Wang, X.: A survey of deep active learning (2020). \doi{10.48550/ARXIV.2009.00236}, \url{https://arxiv.org/abs/2009.00236}

\bibitem{seneractive}
Sener, O., Savarese, S.: Active learning for convolutional neural networks: A core-set approach. In: International Conference on Learning Representations (2018)

\bibitem{Shin_2021_ICCV}
Shin, G., Xie, W., Albanie, S.: All you need are a few pixels: Semantic segmentation with pixelpick. In: Proceedings of the IEEE/CVF International Conference on Computer Vision (ICCV) Workshops. pp. 1687--1697 (October 2021)

\bibitem{Sinha_2019_ICCV}
Sinha, S., Ebrahimi, S., Darrell, T.: Variational adversarial active learning. In: Proceedings of the IEEE/CVF International Conference on Computer Vision (ICCV) (October 2019)

\bibitem{sun2024opdm}
Sun, X., Jiang, H., Savva, M., Chang, A.X.: {OPDMulti}: Openable part detection for multiple objects. In: Proc. of 3D Vision (2024)

\bibitem{tang2022active}
Tang, C., Xie, L., Zhang, G., Zhang, X., Tian, Q., Hu, X.: Active pointly-supervised instance segmentation. In: Computer Vision--ECCV 2022: 17th European Conference, Tel Aviv, Israel, October 23--27, 2022, Proceedings, Part XXVIII. pp. 606--623. Springer (2022)

\bibitem{labelme2016}
Wada, K.: {labelme: Image Polygonal Annotation with Python}. \url{https://github.com/wkentaro/labelme} (2016)

\bibitem{wang2015semantic}
Wang, J., Yuille, A.L.: Semantic part segmentation using compositional model combining shape and appearance. In: Proceedings of the IEEE conference on computer vision and pattern recognition. pp. 1788--1797 (2015)

\bibitem{wang2015joint}
Wang, P., Shen, X., Lin, Z., Cohen, S., Price, B., Yuille, A.L.: Joint object and part segmentation using deep learned potentials. In: Proceedings of the IEEE International Conference on Computer Vision. pp. 1573--1581 (2015)

\bibitem{wang2019shape2motion}
Wang, X., Zhou, B., Shi, Y., Chen, X., Zhao, Q., Xu, K.: Shape2motion: Joint analysis of motion parts and attributes from 3d shapes. In: Proceedings of the IEEE/CVF Conference on Computer Vision and Pattern Recognition. pp. 8876--8884 (2019)

\bibitem{wu2022d}
Wu, T.H., Liou, Y.S., Yuan, S.J., Lee, H.Y., Chen, T.I., Huang, K.C., Hsu, W.H.: D 2 ada: Dynamic density-aware active domain adaptation for semantic segmentation. In: Computer Vision--ECCV 2022: 17th European Conference, Tel Aviv, Israel, October 23--27, 2022, Proceedings, Part XXIX. pp. 449--467. Springer (2022)

\bibitem{xia2017joint}
Xia, F., Wang, P., Chen, X., Yuille, A.L.: Joint multi-person pose estimation and semantic part segmentation. In: Proceedings of the IEEE conference on computer vision and pattern recognition. pp. 6769--6778 (2017)

\bibitem{Xiang_2020_SAPIEN}
Xiang, F., Qin, Y., Mo, K., Xia, Y., Zhu, H., Liu, F., Liu, M., Jiang, H., Yuan, Y., Wang, H., Yi, L., Chang, A.X., Guibas, L.J., Su, H.: {SAPIEN}: A simulated part-based interactive environment. In: The IEEE Conference on Computer Vision and Pattern Recognition (CVPR) (June 2020)

\bibitem{xie2022towards}
Xie, B., Yuan, L., Li, S., Liu, C.H., Cheng, X.: Towards fewer annotations: Active learning via region impurity and prediction uncertainty for domain adaptive semantic segmentation. In: Proceedings of the IEEE/CVF Conference on Computer Vision and Pattern Recognition. pp. 8068--8078 (2022)

\bibitem{Xie_2020_ACCV}
Xie, S., Feng, Z., Chen, Y., Sun, S., Ma, C., Song, M.: Deal: Difficulty-aware active learning for semantic segmentation. In: Proceedings of the Asian Conference on Computer Vision (ACCV) (November 2020)

\bibitem{yan2020rpm}
Yan, Z., Hu, R., Yan, X., Chen, L., Van~Kaick, O., Zhang, H., Huang, H.: Rpm-net: recurrent prediction of motion and parts from point cloud. ACM Transactions on Graphics (TOG)  \textbf{38}(6) (2019)

\bibitem{AL_comp_survey2022}
Zhan, X., Wang, Q., Huang, K.h., Xiong, H., Dou, D., Chan, A.B.: A comparative survey of deep active learning (2022). \doi{10.48550/ARXIV.2203.13450}, \url{https://arxiv.org/abs/2203.13450}

\bibitem{zhou2019continuity}
Zhou, Y., Barnes, C., Lu, J., Yang, J., Li, H.: On the continuity of rotation representations in neural networks. In: Proceedings of the IEEE/CVF Conference on Computer Vision and Pattern Recognition. pp. 5745--5753 (2019)

\bibitem{zhu2020deformable}
Zhu, X., Su, W., Lu, L., Li, B., Wang, X., Dai, J.: Deformable detr: Deformable transformers for end-to-end object detection. In: International Conference on Learning Representations (2020)

\bibitem{zou2023segment}
Zou, X., Yang, J., Zhang, H., Li, F., Li, L., Gao, J., Lee, Y.J.: Segment everything everywhere all at once. arXiv preprint arXiv:2304.06718  (2023)

\end{thebibliography}
